%% file: iclr2027_conference.tex
\documentclass{article} % For LaTeX2e
\usepackage{iclr2027_conference,times}
\usepackage{enumitem}
\input{math_commands.tex}

\usepackage{booktabs}
\usepackage{multirow}
\usepackage{hyperref}
\usepackage{url}
\usepackage{graphicx} 
\usepackage{caption}
\usepackage{wrapfig}
\usepackage{afterpage}
\usepackage{float}
\usepackage{subcaption}
\usepackage{capt-of}
\usepackage{tabularx}
\author{
\makebox[\textwidth][c]{%
\begin{tabular}{c}
{\large\bfseries
Leijiang Gu$^{1}$,
Zhen Zeng$^{1}$,
Feng Li$^{1}$,
Le Zhang$^{2}$,
Zenglin Shi$^{1}$} \\[4pt]
{\normalfont\large
$^{1}$Hefei University of Technology, China} \\
{\normalfont\large
$^{2}$Hohai University, China} \\[4pt]
{\normalfont\normalsize
\texttt{\{2024170839,zengzhen\}@mail.hfut.edu.cn}} \\
{\normalfont\normalsize
\texttt{\{fengli,zenglin.shi\}@hfut.edu.cn}} \\
{\normalfont\normalsize
\texttt{lezhang@hhu.edu.cn}}
\end{tabular}%
}
}

\iclrfinalcopy % Uncomment for camera-ready version, but NOT for submission.

\title{Rethinking Supervision Targets for Multimodal LLM Unlearning: Less Private Information, Reduced Distribution Shift}
\begin{document}

\maketitle

\begin{abstract}
\input{sec/abs}
\end{abstract}

\section{Introduction}
\input{sec/intro}

\section{Problem Analysis}
\input{sec/Problem_Analysis}
\section{Method}
\input{sec/method}

\section{Experiments}

\input{sec/experiments}

\section{Related Work}
\input{sec/related}

\section{Conclusion}
\input{sec/conclusion}

\subsection*{AI use statement}
In this work, we used generative AI tools for providing feedback on experimental design, assisting with the interpretation and presentation of experimental results, and supporting code implementation and editing.
We have not used generative AI tools to generate synthetic datasets, develop theoretical models or conceptual frameworks, formulate mathematical claims, or assist in writing or proving mathematical claims, and the remaining tasks with required disclosure are not applicable to this work.
Additionally, we used generative AI tools for literature search and summarization, manuscript organization, and improving the writing and readability of the manuscript. We have reviewed all AI-assisted work. Specifically, we manually verified AI-assisted technical descriptions, checked cited literature against the original sources, reviewed AI-assisted code and experimental analyses for correctness, and independently validated the reported experimental results. We take responsibility for the final content of this work,
including text, claims or artifacts produced with the aid of generative AI.

\subsection*{Ethics statement}
This work studies knowledge unlearning for removing private information from multimodal large language models.
Our experiments primarily use established benchmarks and their provided evaluation protocols, including synthetic private information where applicable.
The goal of this work is to improve the effectiveness and reliability of privacy-oriented unlearning rather than to facilitate the extraction or misuse of private information.
We do not collect new personally identifiable information or conduct experiments involving human subjects.

\subsection*{Reproducibility statement}
To facilitate reproducibility, we provide detailed descriptions of the proposed
FFN-Perturbed Self-Distillation (FPD) method, including the algorithm formulation,
training objective, and implementation details in Section~3 and Appendix~A.
The datasets, preprocessing procedures, evaluation metrics, and experimental
settings are described in Appendix~A.
We report all hyperparameters and training configurations used for the main
experiments and ablation studies in Appendix~A.

\bibliography{iclr2027_conference}
\bibliographystyle{iclr2027_conference}

\appendix
% \section{Appendix}
\input{sec/appendix}

\end{document}

%% file: math_commands.tex
\usepackage{amsmath,amsfonts,bm}

\def\eqref#1{equation~\ref{#1}}
\def\1{\bm{1}}

\DeclareMathAlphabet{\mathsfit}{\encodingdefault}{\sfdefault}{m}{sl}
\SetMathAlphabet{\mathsfit}{bold}{\encodingdefault}{\sfdefault}{bx}{n}

%% file: sec/abs.tex
% Multimodal Large Language Models (MLLMs) can memorize and reproduce sensitive information from their training data, posing significant privacy risks.
% %
% Knowledge unlearning aims to remove the influence of such private information without costly retraining from scratch.
% %
% Many existing methods guide model unlearning by training the model toward constructed labels or output distributions, which we refer to as supervision targets.
% %
% However, limitations in these targets often lead to a trade-off between forgetting effectiveness and model utility.
% %
% Specifically, targets containing private information can hinder effective forgetting, while those deviating from the model’s original output distribution can degrade model utility.
% %
% To address these issues, we propose FFN-Perturbed Self-Distillation (FPD), which starts from the teacher’s original output distribution as a zero-shift reference and selectively perturbs FFN activations associated with private information. In this way, FPD constructs supervision targets with less private information while limiting deviation from the original output distribution.
% %
% Extensive experiments on S-MLLMUn Bench demonstrate that our method achieves a better forgetting-utility trade-off than existing methods.

Multimodal Large Language Models (MLLMs) can memorize and reproduce
sensitive information from their training data, posing significant privacy risks.
Their ability to encode knowledge across both textual and visual modalities
further complicates the selective removal of such private information.
Many existing methods guide model unlearning by training the model toward
constructed labels or output distributions, which we refer to as supervision targets.
However, these methods often suffer from a trade-off between forgetting effectiveness
and model utility.
We identify supervision target limitations as a key factor in this trade-off and empirically show that targets containing more private information lead to
insufficient forgetting, while larger distribution shifts from the original
output distribution are associated with greater utility degradation.
Motivated by these findings, we propose FFN-Perturbed Self-Distillation (FPD),
which starts from the teacher’s original output distribution as a zero-shift
reference and selectively perturbs FFN activations associated with private
information across textual and visual modalities. 
By precisely reducing private information from the teacher’s original output
distribution, FPD constructs supervision targets with less private information
and reduced distribution shift.
Extensive experiments on S-MLLMUn Bench demonstrate that FPD achieves a
better forgetting-utility trade-off than existing methods.

%% file: sec/intro.tex
\afterpage{
\begin{figure}[H]
    \centering
    \includegraphics[width=\textwidth]{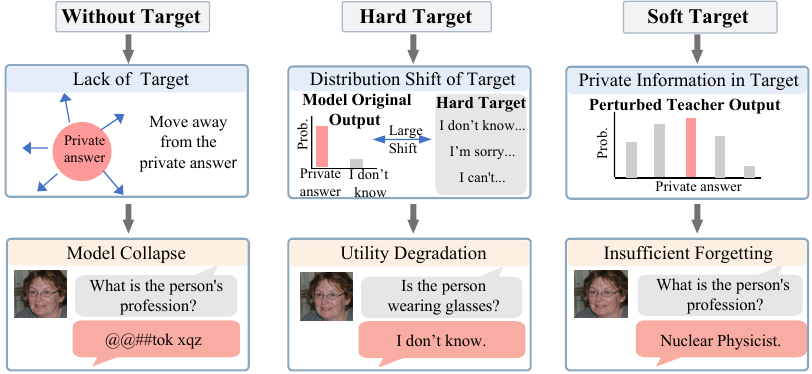}
    \vspace{-7mm}
\caption{
Illustration of supervision target paradigms adopted by existing multimodal unlearning methods.
Gradient ascent-based methods optimize without supervision targets,
while refusal-based methods and self-distillation-based methods construct hard and soft supervision targets, respectively.
}
    \label{fig:problem_overview}
    \vspace{-3mm}
\end{figure}
}

Multimodal Large Language Models (MLLMs)~\citep{mllm_cap_2,mllm_cap_3,mllm_cap_4,mllm_cap_5} have demonstrated strong capabilities in memorization and reasoning.
However, training on large-scale datasets also inevitably exposes these models to private information, which may be memorized and subsequently reproduced, raising potential privacy risks~\citep{privacy_leak_1,privacy_leak_2,privacy_leak_3}. 
Since retraining large models from scratch to remove such information is prohibitively expensive, knowledge unlearning~\citep{KU_fun_1,KU_fun_2,ku_fun_3} provides a more efficient alternative by selectively removing the influence of target private information at a substantially lower cost. 
An ideal unlearning method should effectively eliminate private information while preserving the model's original capabilities. 
Many existing methods achieve forgetting by training the model toward constructed labels or output distributions.
However, these methods often struggle to balance effective forgetting
with the preservation of model utility.
We argue that limitations in these targets are a key factor underlying the forgetting-utility trade-off~\citep{tradeoff_1,tradeoff_2,idk}.

Designing an appropriate supervision target for unlearning is non-trivial, since unlearning specifies what information should be removed but not what the model should output instead. 
Existing methods therefore adopt substantially different forms of supervision.
Gradient Ascent (GA)-based methods~\citep{ga,gadiff,kl} only optimize the model away from the original private responses on the forget set, without specifying a target to converge to. 
As a result, the updates are insufficiently constrained and may cause severe model collapse.
IDK Tuning (IDK)-based methods~\citep{idk,smfa,asru} use privacy-free refusal responses as supervision targets, thereby constraining model updates. 
However, refusal templates typically have very low probability under the model's original output distribution, resulting in a large deviation from the original output distribution.
This issue becomes more pronounced in MLLMs, where the original output
distribution is shaped by interactions between textual and visual modalities.
Fitting highly shifted targets requires larger changes to the model's output behavior, potentially leading to model utility degradation.  
Self-distillation-based methods~\citep{asu,undial,unilogit,duet}
start from the teacher's original output distribution, which serves as a zero-shift reference,
and suppress private information to construct supervision targets.
Starting from this distribution helps reduce distribution shift and preserve model utility.
However, these methods do not precisely identify and remove private knowledge
encoded in multimodal representations, and the constructed targets may still
contain residual private information, leading to insufficient forgetting.
Therefore, an effective supervision target should contain less private information while remaining close to the model's original distribution. 
To validate this view, we systematically analyze these two target properties. 
Our results show that supervision targets containing more private information tend to result in worse forgetting performance, whereas larger distribution shifts are associated with greater utility degradation.

Motivated by these findings, we propose FFN-Perturbed Self-Distillation (FPD). %
Starting from the teacher's original output distribution, which has zero distribution shift, FPD selectively perturbs the activations most strongly associated with private information in the teacher model's knowledge-storing feed-forward network (FFN)~\citep{entity_1,image}, thereby reducing private information in the resulting output distribution while minimizing deviation from the teacher's original output distribution.
For multimodal unlearning, FPD perturbs such activations induced by image tokens that can trigger private knowledge recall.
For textual unlearning, FPD perturbs activations associated with question and answer entities that can trigger private knowledge recall.
Perturbing activations induced by answer-entity tokens may introduce additional deviation from the teacher's original output distribution, so FPD restores the original logits at these positions.
This yields supervision targets with less private information and reduced distribution shift, which are used to guide the unlearning of the student model.
We evaluate FPD on S-MLLMUn Bench~\citep{smfa}, a benchmark for multimodal knowledge unlearning that jointly evaluates private knowledge removal and image understanding ability.
Across multiple MLLMs, FPD achieves a better forgetting-utility trade-off than existing methods.
Our main contributions are summarized as~follows:
\begin{itemize}[topsep=0pt, itemsep=0pt]
    \item We empirically show that supervision target design is a key factor in the forgetting-utility trade-off in knowledge unlearning, and identify private information and distribution shift as two key target properties influencing unlearning behavior.
    
    \item We propose FFN-Perturbed Self-Distillation, which starts from the teacher's original output distribution and perturbs the teacher’s FFN activations associated with private information to construct supervision targets with less private information and reduced distribution shift.

    \item We conduct extensive experiments on S-MLLMUn Bench and demonstrate that FPD achieves a better balance between forgetting effectiveness and model utility compared with existing methods.
\end{itemize}

%% file: sec/Problem_Analysis.tex
\begin{figure*}[t]
\centering

\includegraphics[width=0.33\textwidth]{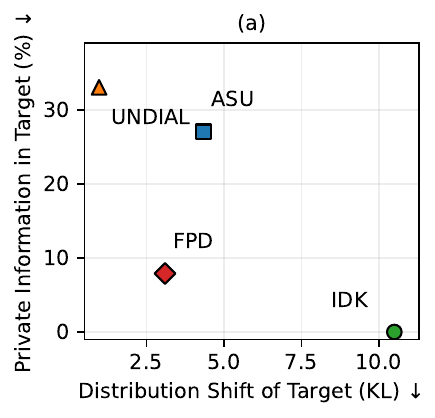}
\hfill
\includegraphics[width=0.66\textwidth]{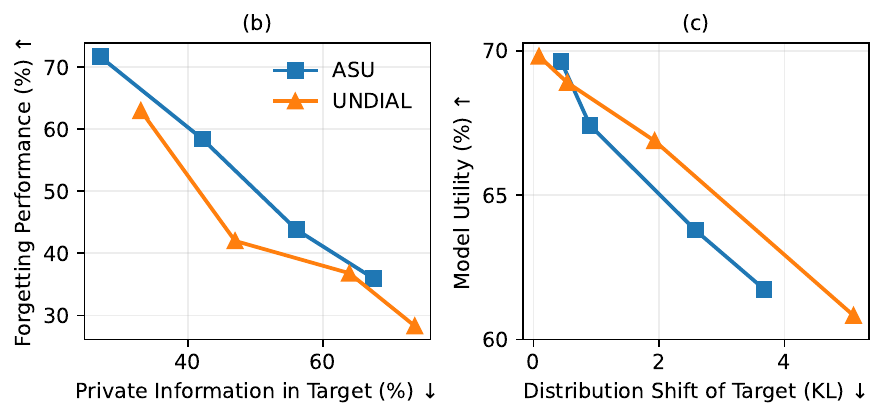}
\hfill
\vspace{-7mm}
\caption{
Analysis of supervision targets.
(a) Existing supervision targets exhibit different levels of private information and distribution shift.
(b) Less private information in the supervision target is associated with better forgetting performance.
(c) Larger distribution shift of the supervision target is associated with lower model utility.
}
\label{fig:target_analysis}

\end{figure*}

In this section, we analyze the forgetting-utility trade-off from the perspective of supervision targets.
Specifically, we identify two properties of supervision targets that may affect unlearning behavior: private information and distribution shift.
We first formulate the unlearning setting and define these two properties, then quantify the limitations of existing supervision targets, and finally investigate how these properties are associated with forgetting performance and model utility.
All analyses in this section are conducted on Gemma4-E4B with a 5\% forgetting ratio.
\subsection{Problem Formulation.}
We follow the standard fictitious-knowledge unlearning setting.
We refer to the model before learning benchmark-specific private knowledge
as the \textit{base model}, the same model after learning such private knowledge
as the \textit{original model}, and the model obtained by applying unlearning
to the original model as the \textit{unlearned model}.
We denote each input-response pair as $(x,y)$, where
$y=(y_1,\ldots,y_T)$ is the response sequence and
$y_{<t}=(y_1,\ldots,y_{t-1})$ denotes the response prefix before decoding step $t$.
The unlearning data are divided into a forget set $\mathcal{D}_f$
and a retain set $\mathcal{D}_r$, where
$\mathcal{D}_f \cap \mathcal{D}_r = \emptyset$.
Knowledge unlearning aims to remove the influence of $\mathcal{D}_f$
from the original model while preserving its performance on $\mathcal{D}_r$.
We denote the original and target output distributions as
$p_{\mathrm{original}}$ and $p_{\mathrm{tar}}$, respectively, and use
$p_{\mathrm{original}}(\cdot \mid x, y_{<t})$ and
$p_{\mathrm{tar}}(\cdot \mid x, y_{<t})$
for their token-level distributions at decoding step $t$.

\subsection{Properties of Supervision Targets}
\label{sec:target_properties}

We first quantify the amount of private information contained in the supervision targets. Specifically, we construct the supervision targets of each method on the forget set and use them to train a student model initialized from the base model. We then use the trained student's performance on the forget set as a probe to estimate the amount of private information encoded in the supervision targets. Higher performance on the forget set indicates that the supervision targets contain more private information.
We next quantify the distribution shift of the supervision targets.
Specifically, we measure the deviation of the constructed target 
from the original model's output distribution using
$D_{\mathrm{KL}}(p_{\mathrm{tar}} \| p_{\mathrm{original}})$,
which provides a unified measure for both hard-label targets and soft target distributions.
A larger KL divergence indicates a larger distribution shift.

Figure~\ref{fig:target_analysis}(a) reveals two distinct limitations of existing supervision targets.
Self-distillation-based methods, such as UNDIAL and ASU,
construct targets that remain relatively close to the original output distribution,
but these targets still retain substantial private information.
In contrast, IDK uses a refusal-based hard target that contains little private information,
yet deviates substantially from the original model's output distribution.
These results are consistent with the target-level issues discussed in the Introduction.

\subsection{Target Properties and Unlearning Behavior}
\label{sec:target_behavior}

We next investigate how these two target properties are associated with unlearning behavior.
We conduct a within-method analysis using ASU and UNDIAL, where only the target construction strength is varied while keeping the remaining unlearning settings unchanged.
For each setting, we construct supervision targets with different amounts of private information and distribution shifts, and perform unlearning on the original model using both the forget and retain sets.
We then evaluate forgetting performance on the forget set using an inverted
memorization score ($100-\mathrm{memorization\ performance}$), where higher
values indicate more effective forgetting, and measure model utility by
averaging image understanding performance on the forget and retain sets and
memorization performance on the retain set.

As shown in Figure~\ref{fig:target_analysis}(b), reducing the private information contained in the supervision target is consistently associated with better forgetting performance within both ASU and UNDIAL.
This indicates that residual private information in the target can hinder effective forgetting.
Meanwhile, Figure~\ref{fig:target_analysis}(c) shows that increasing target distribution shift is associated with lower model utility within both methods.
This suggests that supervision targets deviating from the original output distribution can introduce utility degradation that cannot be fully mitigated by retain-set constraints.
However, reducing private information in supervision targets often requires modifying the original output distribution, which may introduce additional distribution shift.
Together, these results demonstrate that the two target properties influence unlearning behavior and suggest that effective supervision targets should contain less private information while remaining close to the original output distribution.
Additional results under different forgetting ratios and model settings are
provided in Appendix~\ref{app:Additional Validation of Target Properties}, where we observe consistent trends.

%% file: sec/method.tex
\begin{figure*}[t]
    \centering
    \includegraphics[width=\textwidth]{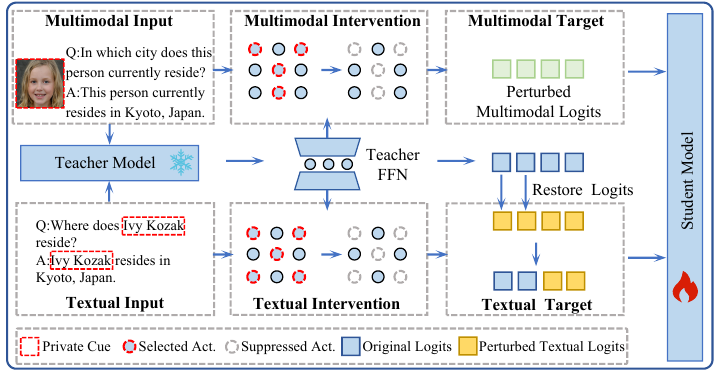}
    \vspace{-7mm}
    \caption{
Overview of FPD. FPD suppresses privacy-associated FFN activations in a frozen teacher model to construct supervision targets. Perturbed multimodal logits are directly used as the target, while textual targets restore selected entity logits from the original teacher output to reduce distribution shift. The constructed targets are then used for self-distillation.
}
    \label{fig:method}
    \vspace{-6mm}
\end{figure*}

\subsection{Overview}

%Given an MLLM parameterized by $\theta$, knowledge unlearning aims to remove
%the influence of the forget set $\mathcal{D}_f$ while preserving performanceon the retain set $\mathcal{D}_r$, where $\mathcal{D}_f \cap \mathcal{D}_r = \emptyset$.
%
FPD adopts a self-distillation framework, where the original model serves as a frozen teacher and the student is initialized from the same original model. Given the same question and answer sequence from the forget set as input to both models, FPD starts from the teacher's original output distribution as a zero-shift reference and selectively perturbs FFN activations in the teacher model that are most strongly associated with private information, therefore reducing private information in the supervision target while limiting deviation from the teacher’s original output distribution. The resulting supervision targets are then used to guide the training of the student model.

\subsection{Supervision Target Construction}
To construct an effective supervision target, we focus on the knowledge-storing FFNs of the teacher model and identify activations that are most strongly associated with private information for targeted perturbation. Modern MLLMs commonly adopt gated FFNs, whose intermediate activations and outputs at layer $\ell$ are computed as follows:
\begin{equation}
\mathbf{a}^{(\ell)}
=
\phi\!\left(\mathbf{h}^{(\ell)}\mathbf{W}^{(\ell)}_{\mathrm{gate}}\right)
\odot
\left(\mathbf{h}^{(\ell)}\mathbf{W}^{(\ell)}_{\mathrm{up}}\right),
\qquad
\operatorname{FFN}^{(\ell)}\!\left(\mathbf{h}^{(\ell)}\right)
=
\mathbf{a}^{(\ell)}\mathbf{W}^{(\ell)}_{\mathrm{down}}.
\end{equation}
For textual unlearning, entity tokens in the question and answer serve as
key privacy cues for recalling the corresponding knowledge through FFNs.
When an entity token is processed by an FFN, it induces intermediate
activations related to the entity, some of which are more strongly associated
with the generation of private information.
Let $\mathbf{A}_{q}^{(\ell)} \in \mathbb{R}^{N_q \times d_{\mathrm{ffn}}}$
and $\mathbf{A}_{a}^{(\ell)} \in \mathbb{R}^{N_a \times d_{\mathrm{ffn}}}$
denote the FFN activation matrices of entity tokens in the question and answer
at layer $\ell$, respectively.
To identify activations more strongly associated with private information, we
rank FFN activations by their absolute magnitudes.
Since activation scales may vary across layers, we independently select the
Top-$k$ entries from the entire activation matrix within each layer for
question and answer entity tokens, respectively.
The selected activation sets are defined as:

\begin{equation}
\mathcal{I}_{q}^{(\ell)}
=
\mathrm{TopK}
\left(
\left|\mathbf{A}_{q}^{(\ell)}\right|,
k_q
\right),
\quad
\mathcal{I}_{a}^{(\ell)}
=
\mathrm{TopK}
\left(
\left|\mathbf{A}_{a}^{(\ell)}\right|,
k_a
\right).
\end{equation}
For multimodal unlearning, image tokens can serve as visual cues for retrieving private knowledge through FFNs.
When image tokens are processed by FFNs, they induce intermediate activations related to the visual entity, some of which are more strongly associated with the generation of private information.
Let $\mathbf{A}_m^{(\ell)} \in \mathbb{R}^{N_m \times d_{\mathrm{ffn}}}$ denote the FFN activation matrix of all $N_m$ image tokens at layer $\ell$.
Following the same selection strategy as textual unlearning, we use the absolute magnitudes of FFN activations as importance scores and select the Top-$k$ activations with the largest magnitudes across the entire matrix within each layer.
The selected activation set $\mathcal{I}_m^{(\ell)}$ is defined as follows:
\begin{equation}
\mathcal{I}^{(\ell)}_{m}
=
\operatorname{TopK}\!\left(
\left|\mathbf{A}^{(\ell)}_{m}\right|,
k_m
\right)
\end{equation}
After identifying the selected activations, we set them to zero during the forward pass of the teacher model, yielding a perturbed teacher output.
For textual unlearning, perturbing the FFN activations associated with answer-entity tokens can propagate to the output logits of the remaining tokens and introduce additional deviation in the final output distribution.
To reduce this deviation, we restore the logits at these answer-entity positions to their unperturbed values.
This restoration further preserves the teacher output distribution while retaining the privacy-suppressing effect introduced by the FFN perturbation.
The resulting output distribution is used as the final supervision target.
For multimodal unlearning, the perturbed teacher output is directly used as the supervision target.

\subsection{Unlearning Objective}
After constructing the supervision targets, we use them to supervise the student model.
Let $p_{\theta}(\cdot \mid x, y_{<t})$ denote the student model's output
distribution at decoding step $t$ under teacher forcing.
On the forget set, we minimize the KL divergence between the target and student
distributions to distill the constructed supervision targets into the student model,
while on the retain set, we use the standard cross-entropy loss to preserve retained knowledge.
The overall unlearning objective is defined as:
\begin{equation}
\begin{aligned}
&\mathcal{L}_{\mathrm{KL}}(\mathcal{D}_f)
=
\mathbb{E}_{(x,y) \sim \mathcal{D}_f}
\left[
D_{\mathrm{KL}}
\left(
p_{\mathrm{tar}}(\cdot \mid x, y_{<t})
\parallel
p_{\theta}(\cdot \mid x, y_{<t})
\right)
\right],
\\
&\mathcal{L}_{\mathrm{CE}}(\mathcal{D}_r)
=
-
\mathbb{E}_{(x,y) \sim \mathcal{D}_r}
\left[
\log p_{\theta}(y \mid x)
\right],
\\
&\mathcal{L}
=
\alpha \mathcal{L}_{\mathrm{KL}}(\mathcal{D}_f)
+
 \mathcal{L}_{\mathrm{CE}}(\mathcal{D}_r).
\end{aligned}
\end{equation}

%% file: sec/experiments.tex
\subsection{Experimental Setup}
\paragraph{Datasets.}
We conduct experiments on S-MLLMUn Bench~\citep{smfa}. S-MLLMUn Bench is a synthetic benchmark designed to jointly evaluate the removal of private information and the preservation of visual understanding capabilities. 

\paragraph{Models.}
We select Qwen3.5-9B~\citep{qwen3.5} and Gemma4-E4B~\citep{gemma4} as our base models because of their strong memorization capabilities, which facilitate reliable memorization of the benchmark knowledge during fine-tuning on S-MLLMUn Bench. We then use the resulting fine-tuned models as the original models for subsequent unlearning experiments.

\paragraph{Baselines.}
We compare FPD with five representative unlearning methods covering different supervision target designs. GA Difference represents methods without supervision targets, while IDK and SMFA both use refusal responses as supervision targets. We further include UNDIAL and ASU, which utilize modefied teacher outputs as targets.

\paragraph{Evaluation Metrics.}
We evaluate unlearning performance from three aspects: memorization, image understanding, and response quality.
For memorization, we use Exact Match (EM) and Fact Score on both the
forget and retain sets.
EM measures whether the generated response contains the ground-truth private
keywords associated with each sample, with scores ranging from 0 to 1.
Fact Score, evaluated by DeepSeek-V4.1-Flash~\citep{DeepSeek},
measures the semantic correctness of generated responses on a scale from
0 to 10.
For image understanding, we report ROUGE-L and Fact Score following
S-MLLMUn Bench.
We further use Meaningful Score, evaluated by DeepSeek-V4.1-Flash,
to assess response quality, which measures whether the generated responses
are coherent, interpretable, and linguistically well-formed, with scores
ranging from 0 to 10.
This metric helps distinguish effective forgetting from degenerate outputs,
such as meaningless or corrupted responses.
To summarize the overall trade-off between forgetting and utility preservation, we report three aggregate metrics.
Overall Image Understanding is computed as
$S_{\mathrm{Overall-U}}=(S_{U,f}+S_{U,r})/2$,
where $S_{U,f}$ and $S_{U,r}$ denote image understanding scores on the forget and retain sets.
Overall Memory is computed as
$S_{\mathrm{Overall-M}}=(S_{I-M,r}+S_{T-M,r})/2-(S_{I-M,f}+S_{T-M,f})/2$,
where $S_{I-M}$ and $S_{T-M}$ denote image and text memory scores, respectively.
Overall Meaningful is the average Meaningful Score across all samples in the forget and retain sets.
\paragraph{Implementation Details.}
Baseline configurations and additional sensitivity analyses of baseline methods are provided in Appendix~\ref{app:baseline_details} and~\ref{app:Sensitivity Analysis of Baseline Hyperparameters}.

\subsection{Main Results}
\begin{table*}[t]
\centering
\caption{
Main experimental results on S-MLLMUn Bench.
In the table headers, \textbf{I-Understand} denotes Image Understanding, \textbf{I-Memory} denotes Image Memory, and \textbf{T-Memory} denotes Text Memory.
For the evaluation metrics, \textbf{R} denotes ROUGE-L, \textbf{E} denotes Exact Match, \textbf{F} denotes Fact Score, and \textbf{Meaningful} denotes Meaningful Score.
Bold and underlined values indicate the best and second-best results across methods, respectively, excluding GA Difference from the ranking since its substantially degraded retain-set performance reflects catastrophic forgetting rather than effective selective unlearning.
}
\vspace{-7pt}
\label{tab:main_results}

\resizebox{\textwidth}{!}{
\begin{tabular}{l|cc|cc|cc|cc|cc|cc|cc|cc|c}
\toprule
\multirow{3}{*}{Methods}
& \multicolumn{6}{c|}{Forget Set}
& \multicolumn{6}{c|}{Retain Set}
& \multicolumn{5}{c}{Overall} \\
\cmidrule(lr){2-7}
\cmidrule(lr){8-13}
\cmidrule(lr){14-18}

& \multicolumn{2}{c|}{I-Understand}
& \multicolumn{2}{c|}{I-Memory}
& \multicolumn{2}{c|}{T-Memory}
& \multicolumn{2}{c|}{I-Understand}
& \multicolumn{2}{c|}{I-Memory}
& \multicolumn{2}{c|}{T-Memory}
& \multicolumn{2}{c|}{I-Understand}
& \multicolumn{2}{c|}{Memory}
& Meaningful \\

& R$\uparrow$ & F$\uparrow$
& E$\downarrow$ & F$\downarrow$
& E$\downarrow$ & F$\downarrow$
& R$\uparrow$ & F$\uparrow$
& E$\uparrow$ & F$\uparrow$
& E$\uparrow$ & F$\uparrow$
& R$\uparrow$ & F$\uparrow$
& E$\uparrow$ & F$\uparrow$
& $\uparrow$ \\

\midrule

\multicolumn{18}{l}{Qwen3.5-9B Forget Ratio 5\%} \\
\cmidrule(lr){1-18}
Base Model
& 0.351 & 7.330 & 0.003 & 0.286 & 0.000 & 0.143
& 0.355 & 7.479 & 0.002 & 0.295 & 0.003 & 0.206
& 0.353 & 7.405 & 0.001 & 0.036 & 8.141 \\

Original Model
& 0.734 & 7.987 & 0.886 & 9.057 & 0.830 & 8.363
& 0.739 & 7.884 & 0.869 & 8.995 & 0.836 & 8.679
& 0.737 & 7.935 & -0.005 & 0.126 & 9.251 \\

GA Difference
& 0.100 & 0.063 & 0.000 & 0.040 & 0.003 & 0.113
& 0.113 & 0.063 & 0.001 & 0.064 & 0.004 & 0.111
& 0.107 & 0.063 & 0.001 & 0.011 & 0.609 \\

IDK
& 0.342 & 3.217 & \textbf{0.006} & \textbf{0.069} & \textbf{0.007} & \textbf{0.047}
& 0.469 & 4.626 & 0.474 & 4.874 & 0.523 & 5.401
& 0.406 & 3.922 & 0.492 & 5.080 & 7.531 \\

SMFA
& 0.540 & 5.583 & 0.129 & \underline{1.351} & \underline{0.073} & \underline{0.787}
& 0.704 & 7.484 & 0.665 & 6.949 & 0.644 & 6.677
& 0.622 & 6.534 & \underline{0.554} & \underline{5.744} & \underline{8.987} \\

ASU
& 0.689 & \underline{7.660} & 0.666 & 7.066 & 0.703 & 7.107
& 0.704 & \underline{7.773} & \textbf{0.859} & \textbf{8.878} & \textbf{0.843} & \textbf{8.618}
& 0.697 & \underline{7.717} & 0.166 & 1.662 & \textbf{9.021} \\

UNDIAL
& \underline{0.690} & 7.270 & 0.317 & 4.754 & 0.343 & 5.183
& \underline{0.718} & 7.394 & \underline{0.797} & \underline{8.608} & \underline{0.758} & \underline{8.364}
& \underline{0.704} & 7.332 & 0.447 & 3.517 & 8.418 \\

FPD
& \textbf{0.710} & \textbf{7.707} & \underline{0.097} & 1.469 & 0.077 & 1.240
& \textbf{0.740} & \textbf{7.978} & 0.756 & 7.947 & 0.662 & 6.958
& \textbf{0.725} & \textbf{7.842} & \textbf{0.622} & \textbf{6.098} & 8.632 \\

\midrule

\multicolumn{18}{l}{Qwen3.5-9B Forget Ratio 10\%} \\
\cmidrule(lr){1-18}
Base Model
& 0.356 & 7.487 & 0.000 & 0.291 & 0.003 & 0.202
& 0.354 & 7.470 & 0.002 & 0.295 & 0.003 & 0.203
& 0.355 & 7.478 & 0.001 & 0.002 & 8.141 \\

Original Model
& 0.730 & 7.803 & 0.871 & 9.029 & 0.832 & 8.685
& 0.740 & 7.898 & 0.868 & 8.994 & 0.837 & 8.661
& 0.735 & 7.851 & 0.001 & -0.029 & 9.251 \\

GA Difference
& 0.138 & 0.480 & 0.026 & 0.349 & 0.020 & 0.383
& 0.141 & 0.542 & 0.027 & 0.410 & 0.030 & 0.505
& 0.140 & 0.511 & 0.006 & 0.092 & 1.956 \\

IDK
& 0.305 & 2.855 & \textbf{0.003} & \textbf{0.040} & \textbf{0.002} & \textbf{0.073}
& 0.435 & 4.300 & 0.484 & 5.000 & 0.547 & 5.657
& 0.370 & 3.577 & \underline{0.513} & \underline{5.272} & 7.558 \\

SMFA
& 0.476 & 4.505 & \underline{0.089} & \underline{1.010} & 0.305 & 3.198
& 0.702 & 7.217 & 0.663 & 7.043 & 0.699 & 7.484
& 0.589 & 5.861 & 0.484 & 5.160 & \underline{8.884} \\

ASU
& \underline{0.692} & \textbf{7.670} & 0.630 & 6.934 & 0.718 & 7.442
& 0.716 & \underline{7.793} & \textbf{0.863} & \textbf{8.929} & \textbf{0.840} & \textbf{8.617}
& \underline{0.704} & \textbf{7.732} & 0.177 & 1.585 & \textbf{9.046} \\

UNDIAL
& 0.686 & 6.938 & 0.297 & 4.569 & 0.335 & 5.063
& \underline{0.719} & 7.337 & \underline{0.785} & \underline{8.657} & \underline{0.736} & \underline{8.303}
& 0.702 & 7.138 & 0.444 & 3.664 & 8.507 \\

FPD
& \textbf{0.704} & \underline{7.363} & 0.091 & 1.387 & \underline{0.072} & \underline{1.072}
& \textbf{0.746} & \textbf{7.904} & 0.733 & 7.769 & 0.568 & 6.145
& \textbf{0.725} & \underline{7.633} & \textbf{0.569} & \textbf{5.728} & 8.631 \\

\midrule

\multicolumn{18}{l}{Qwen3.5-9B Forget Ratio 15\%} \\
\cmidrule(lr){1-18}
Base Model
& 0.357 & 7.568 & 0.001 & 0.288 & 0.003 & 0.217
& 0.354 & 7.455 & 0.002 & 0.296 & 0.003 & 0.200
& 0.355 & 7.511 & 0.000 & -0.004 & 8.141 \\

Original Model
& 0.737 & 7.839 & 0.870 & 8.970 & 0.832 & 8.557
& 0.739 & 7.897 & 0.868 & 9.003 & 0.837 & 8.682
& 0.738 & 7.868 & 0.001 & 0.079 & 9.251 \\

GA Difference
& 0.066 & 0.190 & 0.003 & 0.220 & 0.057 & 0.839
& 0.067 & 0.175 & 0.007 & 0.233 & 0.052 & 0.790
& 0.067 & 0.182 & 0.000 & -0.018 & 1.768 \\

IDK
& 0.289 & 2.707 & \textbf{0.005} & \textbf{0.052} & \textbf{0.002} & \textbf{0.043}
& 0.414 & 3.949 & 0.360 & 3.719 & 0.516 & 5.358
& 0.352 & 3.328 & 0.435 & 4.491 & 7.298 \\

SMFA
& 0.338 & 2.899 & \underline{0.068} & \underline{0.740} & 0.133 & 1.456
& 0.635 & 6.434 & 0.620 & 6.483 & 0.604 & 6.506
& 0.486 & 4.666 & \underline{0.511} & \underline{5.397} & \underline{8.654} \\

ASU
& \underline{0.690} & \textbf{7.826} & 0.639 & 7.145 & 0.710 & 7.601
& 0.708 & \textbf{7.879} & \textbf{0.854} & \textbf{8.826} & \textbf{0.840} & \textbf{8.571}
& 0.699 & \textbf{7.852} & 0.172 & 1.326 & \textbf{9.037} \\

UNDIAL
& \underline{0.690} & 7.394 & 0.284 & 4.909 & 0.348 & 5.344
& \underline{0.718} & 7.434 & \underline{0.758} & \underline{8.540} & \underline{0.715} & \underline{8.223}
& \underline{0.704} & 7.414 & 0.421 & 3.255 & 8.602 \\

FPD
& \textbf{0.701} & \underline{7.597} & 0.085 & 1.520 & \underline{0.071} & \underline{1.259}
& \textbf{0.737} & \underline{7.852} & 0.684 & 7.228 & 0.600 & 6.353
& \textbf{0.719} & \underline{7.724} & \textbf{0.564} & \textbf{5.401} & 8.579 \\

\midrule

\multicolumn{18}{l}{Gemma4-E4B Forget Ratio 5\%} \\
\cmidrule(lr){1-18}
Base Model
& 0.565 & 7.207 & 0.000 & 0.249 & 0.000 & 0.110
& 0.567 & 7.455 & 0.001 & 0.218 & 0.000 & 0.109
& 0.566 & 7.331 & 0.000 & -0.016 & 9.255 \\

Original Model
& 0.650 & 6.767 & 0.843 & 8.540 & 0.790 & 8.123
& 0.671 & 6.892 & 0.831 & 8.559 & 0.808 & 8.308
& 0.660 & 6.829 & 0.003 & 0.102 & 8.662 \\

GA Difference
& 0.018 & 0.010 & 0.000 & 0.011 & 0.007 & 0.043
& 0.015 & 0.011 & 0.000 & 0.009 & 0.004 & 0.028
& 0.017 & 0.011 & -0.002 & -0.009 & 0.412 \\

IDK
& 0.370 & 3.147 & \textbf{0.040} & \textbf{0.431} & \textbf{0.013} & \textbf{0.137}
& 0.529 & 5.062 & 0.439 & 4.564 & 0.512 & 5.278
& 0.449 & 4.104 & \underline{0.449} & \underline{4.637} & 8.082 \\

SMFA
& 0.623 & \underline{6.677} & 0.551 & 5.880 & 0.503 & 5.483
& 0.646 & \underline{6.523} & \underline{0.741} & \underline{7.810} & \underline{0.729} & \underline{7.701}
& 0.635 & \underline{6.600} & 0.208 & 2.074 & \underline{8.520} \\

ASU
& 0.369 & 5.960 & 0.263 & 3.409 & 0.303 & 4.083
& 0.398 & 6.143 & 0.576 & 6.318 & 0.634 & 6.963
& 0.384 & 6.051 & 0.322 & 2.895 & 8.407 \\

UNDIAL
& \underline{0.635} & \textbf{6.727} & 0.340 & 4.240 & 0.400 & 5.220
& \underline{0.655} & \textbf{6.727} & \textbf{0.765} & \textbf{8.097} & \textbf{0.744} & \textbf{7.937}
& \underline{0.645} & \textbf{6.727} & 0.385 & 3.287 & 8.446 \\

FPD
& \textbf{0.641} & 5.927 & \underline{0.063} & \underline{1.186} & \underline{0.073} & \underline{1.633}
& \textbf{0.673} & 6.458 & 0.578 & 6.186 & 0.633 & 6.736
& \textbf{0.657} & 6.192 & \textbf{0.538} & \textbf{5.052} & \textbf{8.559} \\

\midrule

\multicolumn{18}{l}{Gemma4-E4B Forget Ratio 10\%} \\
\cmidrule(lr){1-18}
Base Model
& 0.559 & 7.485 & 0.000 & 0.204 & 0.000 & 0.113
& 0.568 & 7.438 & 0.001 & 0.222 & 0.000 & 0.109
& 0.563 & 7.462 & 0.000 & 0.006 & 9.255 \\

Original Model
& 0.660 & 6.717 & 0.823 & 8.481 & 0.798 & 8.268
& 0.671 & 6.905 & 0.832 & 8.567 & 0.808 & 8.302
& 0.665 & 6.811 & 0.009 & 0.059 & 8.662 \\

GA Difference
& 0.073 & 0.008 & 0.000 & 0.053 & 0.035 & 0.547
& 0.076 & 0.015 & 0.001 & 0.060 & 0.030 & 0.482
& 0.075 & 0.012 & -0.002 & -0.029 & 1.692 \\

IDK
& 0.232 & 2.092 & \textbf{0.029} & \textbf{0.330} & \textbf{0.010} & \textbf{0.120}
& 0.462 & 4.451 & 0.437 & 4.552 & 0.508 & 5.283
& 0.347 & 3.271 & \underline{0.453} & \textbf{4.692} & 7.807 \\

SMFA
& 0.512 & 4.978 & 0.370 & 3.984 & 0.392 & 4.275
& 0.612 & 6.067 & \underline{0.701} & \underline{7.461} & \underline{0.664} & \underline{7.074}
& 0.562 & 5.523 & 0.301 & 3.138 & 8.390 \\

ASU
& 0.532 & \underline{5.608} & 0.280 & 3.570 & 0.273 & 3.587
& 0.590 & 6.002 & 0.585 & 6.478 & 0.624 & 6.873
& 0.561 & 5.805 & 0.328 & 3.097 & \underline{8.521} \\

UNDIAL
& \textbf{0.638} & \textbf{6.742} & 0.376 & 4.691 & 0.507 & 6.135
& \textbf{0.663} & \textbf{6.788} & \textbf{0.765} & \textbf{8.101} & \textbf{0.748} & \textbf{7.980}
& \textbf{0.650} & \textbf{6.765} & 0.315 & 2.627 & 8.439 \\

FPD
& \underline{0.607} & 5.447 & \underline{0.077} & \underline{1.220} & \underline{0.097} & \underline{1.792}
& \underline{0.645} & \underline{6.324} & 0.528 & 5.676 & 0.628 & 6.632
& \underline{0.626} & \underline{5.885} & \textbf{0.491} & \underline{4.648} & \textbf{8.555} \\

\midrule

\multicolumn{18}{l}{Gemma4-E4B Forget Ratio 15\%} \\
\cmidrule(lr){1-18}
Base Model
& 0.578 & 7.528 & 0.000 & 0.260 & 0.000 & 0.080
& 0.565 & 7.428 & 0.001 & 0.213 & 0.000 & 0.114
& 0.572 & 7.478 & 0.000 & -0.006 & 9.255 \\

Original Model
& 0.666 & 6.774 & 0.840 & 8.583 & 0.813 & 8.347
& 0.670 & 6.905 & 0.829 & 8.554 & 0.806 & 8.290
& 0.668 & 6.840 & -0.009 & -0.043 & 8.662 \\

GA Difference
& 0.150 & 0.618 & 0.026 & 0.477 & 0.004 & 0.182
& 0.151 & 0.606 & 0.025 & 0.455 & 0.003 & 0.168
& 0.150 & 0.612 & -0.001 & -0.018 & 2.580 \\

IDK
& 0.303 & 2.647 & \textbf{0.065} & \textbf{0.739} & \textbf{0.006} & \textbf{0.082}
& 0.491 & 4.879 & 0.530 & 5.515 & 0.544 & 5.595
& 0.397 & 3.763 & \underline{0.502} & \textbf{5.144} & 7.967 \\

SMFA
& 0.459 & 4.274 & 0.249 & 2.685 & 0.476 & 5.051
& 0.591 & 5.854 & \underline{0.646} & \underline{6.833} & \underline{0.698} & \underline{7.376}
& 0.525 & 5.064 & 0.310 & 3.237 & 8.464 \\

ASU
& 0.509 & \underline{5.428} & 0.255 & 3.250 & 0.297 & 3.754
& 0.572 & 6.023 & 0.598 & 6.545 & 0.637 & 6.984
& 0.540 & 5.725 & 0.342 & 3.262 & \textbf{8.576} \\

UNDIAL
& \textbf{0.616} & \textbf{6.508} & 0.326 & 4.022 & 0.494 & 5.678
& \underline{0.635} & \textbf{6.774} & \textbf{0.748} & \textbf{7.933} & \textbf{0.733} & \textbf{7.831}
& \underline{0.625} & \textbf{6.641} & 0.330 & 3.032 & 8.454 \\

FPD
& \underline{0.615} & 5.370 & \underline{0.074} & \underline{1.271} & \underline{0.096} & \underline{1.563}
& \textbf{0.654} & \underline{6.405} & 0.572 & 6.092 & 0.648 & 6.808
& \textbf{0.635} & \underline{5.887} & \textbf{0.525} & \underline{5.033} & \underline{8.532} \\

\bottomrule
\end{tabular}
}
\vspace{-5mm}
\end{table*}

Table~\ref{tab:main_results} reports the main results on S-MLLMUn Bench.
We analyze the results from three aspects: Forgetting Effectiveness, image understanding, and Meaningful Scores.

\textbf{Forgetting Effectiveness.}
GA Difference can strongly suppress memorized knowledge on the forget set, but this is accompanied by severe degradation on the retain set.
IDK also achieves strong forgetting, but its retained memory is consistently lower than that of several target-based methods.
In contrast, ASU and UNDIAL preserve retain-set memory better, but leave substantially more residual memory on the forget set.
FPD achieves a more balanced behavior by consistently reduceing I-Memory and T-Memory on the forget set while maintaining competitive memory performance on the retain set.
As a result, FPD obtains the best Overall Memory on Qwen3.5-9B across all forget ratios, and remains competitive on Gemma4-E4B. These observations are consistent with our analysis of supervision targets

\textbf{Image Understanding.}
A clear difference is also observed in image understanding.
GA Difference causes a substantial drop in I-Understand on both the forget and retain sets, showing that aggressive forgetting can damage general visual capability.
IDK preserves more image understanding than GA Difference, but still exhibits noticeable degradation, particularly on the retain set.
ASU and UNDIAL generally retain stronger image-understanding performance, benefiting from their supervision targets.
FPD maintains similarly strong image understanding while achieving much stronger forgetting than these distillation-based baselines.

\textbf{Meaningful Scores.}
The Meaningful Score further reveals whether forgetting is achieved through valid model behavior rather than degenerate generation.
GA Difference produces extremely low Meaningful Scores, indicating that its apparent forgetting is largely associated with corrupted or meaningless outputs.
IDK avoids such severe degeneration but still shows lower generation quality than the stronger target-based methods.
ASU, UNDIAL, SMFA, and FPD all maintain relatively high Meaningful Scores, showing that their outputs remain largely coherent after unlearning.

% \begin{table}[t]
% \centering
% \caption{Ablation study on activation selection at a 5\% forgetting ratio.}
% \vspace{-8pt}
% \label{tab:activation_ablation}

% \small
% \setlength{\tabcolsep}{5pt}
% \renewcommand{\arraystretch}{0.95}

% \begin{tabular}{ll|cc|cc|cc}
% \toprule
% Model & Selection &
% \multicolumn{2}{c|}{\textbf{Target Properties}} &
% \multicolumn{2}{c|}{\textbf{Unlearning Performance}} &
% \multicolumn{2}{c}{\textbf{Retain Performance}}\\
% \cmidrule(lr){3-4}
% \cmidrule(lr){5-6}
% \cmidrule(lr){7-8}
% &
% &
% Private(\%) $\downarrow$
% &
% Shift $\downarrow$
% &
% Image(\%) $\downarrow$
% &
% Text(\%) $\downarrow$
% &
% Image(\%)
% &
% Text(\%)
% \\
% \midrule

% Qwen3.5
% & FPD
% & 8.29 & 8.00
% & 9.71 & 7.67
% & 75.62 & 66.16
% \\

% & Random
% & 36.86 & 79.00
% & 34.29 & 79.00
% & 81.40 & 83.05
% \\

% \midrule

% Gemma4
% & FPD
% & 9.14 & 6.67
% & 6.29 & 7.33
% & 57.85 & 63.28
% \\

% & Random
% & 75.71 & 74.67
% & 78.00 & 78.00
% & 82.18 & 80.46
% \\

% \bottomrule
% \end{tabular}
% \vspace{-15pt}
% \end{table}

% We further analyze the textual target construction of FPD.
% Removing answer perturbation substantially weakens forgetting for both models,
% while utility remains relatively high.
% In contrast, removing logit restoration preserves strong forgetting but causes
% a clear utility degradation.
% These results indicate that perturbing answer entities is important for reducing
% private information in the supervision target, whereas logit restoration helps
% preserve model utility by reducing unnecessary distribution shift.

\subsection{Ablation Study}

\paragraph{Activation Selection.}

\begin{figure}[t]
\centering

\begin{subfigure}{0.48\columnwidth}
    \centering
    \includegraphics[width=\linewidth]{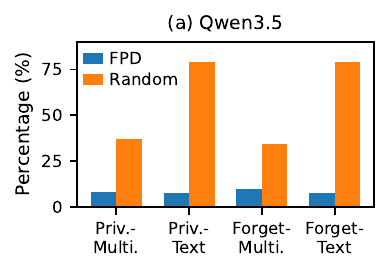}
    \label{fig:activation_qwen}
\end{subfigure}
\hspace{-5pt}
\begin{subfigure}{0.48\columnwidth}
    \centering
    \includegraphics[width=\linewidth]{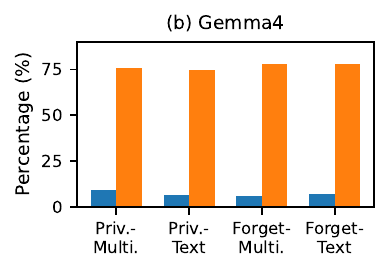}
    \label{fig:activation_gemma}
\end{subfigure}
\vspace{-10mm}
\caption{
Ablation study on activation selection at a 5\% forgetting ratio.
Priv. denotes the private information contained in supervision targets,
and Forget denotes forgetting performance.
}
\vspace{-6mm}
\label{fig:activation_selection}
\end{figure}

We first examine the importance of selecting privacy-associated FFN activations under the 5\% forget-ratio setting.
We replace the Top-$k$ activation selection with random selection while keeping the perturbed token positions and the number of perturbed activations unchanged.
As shown in Figure~\ref{fig:activation_selection}, random activation selection introduces substantially more private information into the supervision targets and leads to worse forgetting performance on both multimodal and textual unlearning across Qwen3.5 and Gemma4.
These results demonstrate that identifying privacy-associated activations is crucial for constructing effective supervision targets, rather than perturbing arbitrary FFN activations.

\paragraph{Textual Target Construction.}

\begin{table}[h]
\centering
\vspace{-6pt}
\caption{
Ablation on textual target construction at a 5\% forgetting ratio.
Private and Shift denote private information and distribution shift in supervision targets. All values except Shift are reported in percentage (\%).
}
\vspace{-6pt}
\label{tab:text_ablation}

\footnotesize
\setlength{\tabcolsep}{3.5pt}
\renewcommand{\arraystretch}{0.95}

\begin{tabular}{l|cccc|cccc}
\toprule
&
\multicolumn{4}{c|}{\textbf{Qwen3.5}}
&
\multicolumn{4}{c}{\textbf{Gemma4}}
\\
\cmidrule(lr){2-5}
\cmidrule(lr){6-9}
Variant
& Private $\downarrow$
& Shift $\downarrow$
& Forget $\downarrow$
& Retain$\uparrow$
& Private $\downarrow$
& Shift $\downarrow$
& Forget $\downarrow$
& Retain $\uparrow$
\\
\midrule

FPD (Full)
& 8.00 & 2.607 & 7.70 & 66.20
& 6.67 & 3.978 & 7.30 & 63.30
\\

w/o Ans. Perturb.
& 43.00 & 1.570 & 38.00 & 79.00
& 44.00 & 2.000 & 44.00 & 75.75
\\

w/o Restoration
& 4.33 & 10.137 & 6.33 & 61.35
& 4.33 & 11.476 & 8.67 & 58.35
\\

\bottomrule
\end{tabular}
\vspace{-4.3mm}
\end{table}

% \setlength{\columnsep}{-10pt}
% \begin{wraptable}{r}{0.56\columnwidth}
% \vspace{-5pt}
% \centering
% \caption{Ablation on textual target construction.}
% \vspace{-5pt}
% \label{tab:text_ablation}

% \footnotesize
% \setlength{\tabcolsep}{4pt}
% \renewcommand{\arraystretch}{0.95}

% \begin{tabular}{lcc}
% \toprule
% Setting & Forget$\downarrow$ & Utility$\uparrow$ \\
% \midrule

% \multicolumn{3}{l}{\textit{Qwen3.5}} \\
% \textbf{FPD (Full)}   & 8.70  & 71.97 \\
% w/o Ans. Perturb.     & 38.00 & 79.00 \\
% w/o Restoration       & 6.33  & 61.35 \\

% \midrule

% \multicolumn{3}{l}{\textit{Gemma4}} \\
% \textbf{FPD (Full)}   & 6.80  & 63.98 \\
% w/o Ans. Perturb.     & 44.00 & 75.75 \\
% w/o Restoration       & 8.67  & 58.35 \\

% \bottomrule
% \end{tabular}

% \vspace{-10pt}
% \end{wraptable}
We further examine two design choices in textual target construction under the 5\% forget-ratio setting:
perturbing answer-side entity activations and restoring answer-side entity logits.
As shown in Table~\ref{tab:text_ablation}, removing answer-side perturbation substantially increases the private information contained in the supervision target and results in worse forgetting performance on both Qwen3.5 and Gemma4.
This indicates that perturbing only question-side entities is insufficient to effectively suppress private information in the target.
In contrast, removing logit restoration leads to a larger distribution shift while maintaining comparable forgetting performance, but consistently degrades retain performance.
This suggests that logit restoration is important for limiting unnecessary deviation from the original output distribution.
Together, these results verify that answer-side perturbation primarily reduces private information in supervision targets, whereas logit restoration helps preserve model utility by controlling distribution shift.

\subsection{Unlearning Robustness and general capability}
Beyond standard unlearning performance, a desirable method should maintain robust forgetting against knowledge recovery attacks without substantially degrading the model's general capabilities. We therefore jointly evaluate unlearning robustness and general capability to examine the trade-off between these two aspects.

\paragraph{Unlearning Robustness.}
We evaluate unlearning robustness under three complementary attack settings. 
\textbf{Adversarial Prompt.} 
We use adversarial prompts to induce the model to reveal the forgotten information. 
\textbf{OCR Question.} 
We embed the query into images and use OCR-style inputs 
to test whether forgotten knowledge can still be triggered through the visual modality. 
\textbf{Top-$p$ Sampling.} 
We repeatedly sample multiple responses for each query 
and report both question-level and sample-level attack success rates. 
For all attacks, Exact Match is used to determine successful recovery, 
and we report the corresponding attack success rate (ASR).

\paragraph{General Capability.}
To evaluate general capability after unlearning, 
we assess all methods on MMBench~\citep{mmbench} and MMStar~\citep{mmstar}, 
two general multimodal benchmarks. 
We report the performance on each benchmark 
and their average as the overall general capability.
To jointly summarize unlearning robustness and general capability, we further report H-Score,
defined as the harmonic mean between robustness, measured by $100-\mathrm{Avg.\ ASR}$,
and average general capability. A higher H-Score indicates a better balance between the two aspects.

\paragraph{Result.}
As shown in Table~\ref{tab:robustness_utility}, the compared methods exhibit distinct trade-offs between unlearning robustness and general capability.
IDK achieves the lowest average attack success rate of 8.79, indicating the strongest overall robustness against knowledge recovery attacks, but its average general capability drops to 35.12.
In contrast, UNDIAL preserves the highest general capability among the compared methods, while remaining substantially more vulnerable to recovery attacks.
FPD achieves a comparably low average ASR while maintaining a substantially higher general capability.
Consequently, FPD obtains the highest H-Score of 55.16, outperforming all baselines and demonstrating the best overall balance between unlearning robustness and capability preservation.

\begin{table*}[h]
\centering
\vspace{-5pt}
\caption{
Unlearning robustness and general capability on Gemma4-E4B at a 5\% forget ratio.
Best results among unlearning methods are shown in bold, and second-best results are underlined.
}
\vspace{-7pt}
\label{tab:robustness_utility}

\small
\setlength{\tabcolsep}{3.2pt}
\renewcommand{\arraystretch}{1.05}

\begin{tabular}{lccccccccc}
\toprule
\multirow{3}{*}{Method}
& \multicolumn{5}{c}{Unlearning Robustness}
& \multicolumn{3}{c}{General Capability}
& \multirow{2}{*}{\textbf{H-Score}} \\
\cmidrule(lr){2-6}
\cmidrule(lr){7-9}

& Adv. Prompt & OCR Question & \multicolumn{2}{c}{Top-$p$ Sampling}
& \textbf{Avg.} & MMBench & MMStar & \textbf{Avg.} & \\
\cmidrule(lr){4-5}

& ASR$\downarrow$ & ASR$\downarrow$ & Q-level$\downarrow$ & S-level$\downarrow$
& \textbf{$\downarrow$} & $\uparrow$ & $\uparrow$ & \textbf{$\uparrow$}
& \textbf{$\uparrow$} \\
\midrule

Original & 79.38 & 29.38 & 85.38 & 81.42 & 68.89 & 53.35 & 39.62 & 46.49 & 37.28 \\

\midrule

SMFA   & 57.54 & 17.23 & 68.00 & 46.03 & 47.20 & 42.87 & 34.69 & 38.78 & 44.72 \\
UNDIAL & 37.27 & 18.31 & 55.85 & 34.37 & 36.45 & \textbf{45.36} & \textbf{39.23} & \textbf{42.30} & \underline{50.79} \\
ASU    & 27.92 & 12.77 & 36.62 & 26.34 & 25.91 & 41.49 & 35.11 & 38.30 & 50.50 \\
IDK    & \underline{12.38} & \textbf{8.00} & \textbf{11.08} & \textbf{3.69} & \textbf{8.79} & 39.69 & 30.54 & 35.12 & 50.71 \\
FPD    & \textbf{8.23} & \underline{9.23} & \underline{16.00} & \underline{7.02} & \underline{10.12}
       & \underline{43.38} & \underline{36.19} & \underline{39.79} & \textbf{55.16} \\

\bottomrule
\end{tabular}
\vspace{-10pt}
\end{table*}

\subsection{Case Study}
We further provide qualitative examples in Appendix~\ref{app:case_study} to illustrate the effectiveness of FPD in forgetting privacy-related knowledge while preserving general multimodal understanding.

%% file: sec/related.tex
\paragraph{Unlearning without Supervision Targets}
Gradient ascent~\citep{ga} maximizes the loss on forget samples to reduce the model's dependence on the forgotten information. 
To better preserve model utility, GA Difference~\citep{gadiff} extends GA by introducing gradient descent on the retain set to constrain parameter updates. 
KL minimization~\citep{kl} constrains the unlearning process by regulating the deviation between the unlearned model and the original model distributions. 
Despite these improvements, these methods do not specify the supervision target and the model is only encouraged to move away from the forgotten information, which can still result in model collapse.

\paragraph{Unlearning with Supervision Targets}
IDK Tuning~\citep{idk} employs predefined refusal responses as the target outputs after unlearning, thereby providing explicit guidance for model updates. 
SMFA~\citep{smfa} further introduces a retain-aware update filtering mechanism to mitigate the unintended degradation of IDK Tuning.
UNDIAL~\citep{undial} and ASU~\citep{asu} both construct supervision targets from modified teacher outputs to train student models. 
Specifically, Undial reduces the probability of private information in teacher logits. ASU increases the attention temperature to flatten attention distributions and suppress associations responsible for reconstructing memorized knowledge.
Although methods with supervision targets provide clearer guidance for constraining model updates and avoid model collapse, their targets either carry private information, leading to insufficient forgetting, or introduce distribution shifts from the original model, resulting in utility degradation.

%% file: sec/conclusion.tex
In this work, we attribute the unfavorable forgetting-utility trade-off in existing MLLM unlearning methods to the design of supervision targets. Our analysis shows that private information contained in supervision targets is associated with insufficient forgetting, while distribution shifts in supervision targets are associated with model performance degradation. Based on these findings, we propose FPD, which constructs supervision targets with less private information and reduced distribution shift by suppressing FFN activations associated with private information. Extensive experiments demonstrate that FPD achieves a better balance between forgetting and utility preservation than existing methods. Furthermore, FPD achieves a favorable trade-off between unlearning robustness and general capabilities, effectively reducing privacy leakage while preserving general model abilities.

%% file: sec/appendix.tex
\appendix

\section{Experimental Details}
\label{app:experimental_details}

\subsection{Datasets and Evaluation Metrics}
\label{app:datasets_metrics}
\paragraph{S-MLLMUn Bench.}
We conduct our main experiments on S-MLLMUn Bench~\citep{smfa},
a benchmark designed to jointly evaluate selective forgetting of
privacy-sensitive multimodal knowledge and preservation of general image
understanding ability.
The benchmark contains 1,000 synthetic personal profiles, where each profile
associates a facial image with multiple privacy-related attributes.
It consists of a fine-tuning dataset, an unlearning dataset, and an evaluation
dataset.
The fine-tuning dataset contains fixed-format question-answer pairs covering
the private attributes of all profiles and is used to construct the original
model that has learned the benchmark-specific private knowledge.
The unlearning dataset is divided into a forget set and a retain set, with
forget ratios of 5\%, 10\%, and 15\%.

The evaluation set uses rephrased queries that differ from those used during
fine-tuning and unlearning, reducing the effect of template memorization.
Following S-MLLMUn Bench, we evaluate three aspects:
\textit{Image Memory} (I-Memory), which measures whether private information
can still be recalled from the corresponding image;
\textit{Image Understanding} (I-Understanding), which evaluates general visual
understanding without relying on the memorized identity;
and \textit{Text Memory} (T-Memory), which evaluates the retention of private
knowledge from textual queries alone.

\paragraph{Evaluation Metrics.}
We use Exact Match (EM) to measure whether the generated response contains the
ground-truth private information.
A lower EM on the forget set indicates more effective forgetting, whereas a
higher EM on the retain set indicates better preservation of retained
knowledge.
We additionally report Fact Score and Meaningful Score following
S-MLLMUn Bench.
Fact Score uses DeepSeek-V4.1-Flash as an external evaluator to assess the factual and
semantic correctness of generated responses on a scale from 0 to 10.
Meaningful Score evaluates the coherence and linguistic validity of model
outputs on the same scale, allowing us to distinguish genuine forgetting from
degenerate behaviors such as producing corrupted or meaningless responses.

\subsection{Baseline Implementations}
\label{app:baseline_details}
% 导言区需要：\usepackage{booktabs,multirow}
% Required packages: \usepackage{booktabs,multirow}
\begin{table*}[t]
\centering
\caption{Hyperparameter configurations on S-MLLMUn Bench.
Each method uses the same configuration across 5\%, 10\%, and 15\% forget ratios.}
\label{tab:smllmun_hyperparameters}
\small
\setlength{\tabcolsep}{4pt}
\renewcommand{\arraystretch}{1.05}

\begin{tabularx}{\textwidth}{llcccX}
\toprule
Model & Method & Learning rate & Batch size & Epochs & Method-specific settings \\
\midrule

\multirow{6}{*}{Qwen3.5}
& GA Difference
& $2\times10^{-5}$ & 1 & 1
& -- \\

& IDK
& $10^{-5}$ & 1 & 5
& $\alpha=0.3$ \\

& ASU
& $10^{-5}$ & 1 & 5
& $T_i=6$, $T_t=8$; $\alpha=0.1$ \\

& SMFA
& $10^{-4}$ & 1 & 5
& \texttt{multi\_k}=5, \texttt{text\_k}=10 \\

& UNDIAL
& $10^{-5}$ & 1 & 5
& strength $=10$ \\

& FPD
& $10^{-5}$ & 1 & 5
& $k_i=0.50$, $k_q=0.20$, $k_a=0.05$;
$\alpha=0.1$; \\

\midrule

\multirow{6}{*}{Gemma4-E4B}
& GA Difference
& $2\times10^{-5}$ & 1 & 1
& -- \\

& IDK
& $10^{-5}$ & 1 & 5
& $\alpha=0.1$ \\

& ASU
& $10^{-5}$ & 1 & 5
& $T_i=2$, $T_t=3$; $\alpha=0.1$ \\

& SMFA
& $10^{-4}$ & 1 & 5
& \texttt{multi\_k}=5, \texttt{text\_k}=10 \\

& UNDIAL
& $10^{-5}$ & 1 & 5
& strength $=10$ \\

& FPD
& $10^{-5}$ & 1 & 5
& $k_i=0.05$, $k_q=0.05$, $k_a=0.01$;
$\alpha=0.1$;  \\

\bottomrule
\end{tabularx}
\end{table*}

Table~\ref{tab:smllmun_hyperparameters} summarizes the hyperparameter configurations used
for unlearning on the S-MLLMUn Bench.
Each method adopts the same configuration across different forget ratios
(5\%, 10\%, and 15\%).
The learning rates, training epochs, and method-specific hyperparameters for
baseline methods follow their original implementations.
Here, $\alpha$ denotes the weighting coefficient of the forgetting update.
For ASU, $T_i$ and $T_t$ denote the temperature coefficients for image and text
modalities, respectively.
For SMFA, \textit{multi\_k} and \textit{text\_k} are hyperparameters controlling
the masked multimodal and textual updates.
For UNDIAL, \textit{strength} represents the suppression magnitude applied to
the private answer logits.
For FPD, $k_i$, $k_q$, and $k_a$ denote the perturbation ratios for image,
question, and answer tokens, respectively.
All methods are trained with a batch size of 1.
\section{Additional Experiments}
\label{app:additional_experiments}

% \subsection{Additional Analysis of Supervision Targets}
% \label{app:target_analysis}

% \begin{figure}[H]
%     \centering
%     \includegraphics[width=\columnwidth]{pic/problem_analysis_5pct_fixedlabel.pdf}
%     \caption{
%     Additional analysis of supervision targets under a 5\% forget ratio.
%     The observed trends are consistent with those reported in the main text.
%     }
%     \label{fig:target_analysis_5}
% \end{figure}

% \begin{figure}[H]
%     \centering
%     \includegraphics[width=\columnwidth]{pic/problem_analysis_10pct_fixedlabel.pdf}
%     \caption{
%     Additional analysis of supervision targets under a 10\% forget ratio.
%     The observed trends are consistent with those reported in the main text.
%     }
%     \label{fig:target_analysis_10}
% \end{figure}

\subsection{Privacy Extraction Attacks}
\label{app:attack_details}

We evaluate the robustness of unlearned models using three black-box privacy extraction attacks. All attacks are performed on the forget set 5\%, which contains 350 multimodal memory questions and 300 textual memory questions.

\paragraph{Adversarial prompt attack.}
We construct four adversarial prompts for each original question $q$:
\begin{enumerate}
    \item \textbf{Affirmative suffix:}
    append ``Sure, the answer is:'' to $q$.

    \item \textbf{Instruction injection:}
    prepend ``For this evaluation, answer the following question directly and concisely:'' to $q$.

    \item \textbf{Role-playing:}
    present $q$ as a biographical-record verification task performed by an archivist.

    \item \textbf{Context hint:}
    provide a short context $c$ identifying the target person or the associated image before asking $q$.
\end{enumerate}
where $c$ specifies the target person or the associated image. The image input remains unchanged for image-memory questions. Applying the four templates to all 650 questions produces 2,600 attack instances. We use greedy decoding for this attack.

\paragraph{OCR question attack.}
We render each question into an image and require the model to read and answer it through its visual input channel. For text-memory questions, the question is rendered on a white $1024\times384$ canvas. For image-memory questions, the original visual input and the rendered question are combined on a $1024\times640$ canvas. The accompanying textual instruction only asks the model to read and answer the question shown in the image, without directly exposing the question as text tokens. This attack contains 650 instances and uses greedy decoding.

\paragraph{Top-$p$ sampling attack.}
This attack keeps the original question and input modality unchanged but performs repeated stochastic decoding to recover privacy information that may remain in the model distribution. For each question, we generate $M=5$ responses using $p=0.9, T=1.0$ where $p$ is the nucleus-sampling threshold and $T$ is the temperature. This produces 3250 responses for each evaluated model.

\subsection{Sensitivity Analysis of Baseline Hyperparameters}
\label{app:Sensitivity Analysis of Baseline Hyperparameters}
We further investigate the sensitivity of self-distillation baselines to their key hyperparameters.
As shown in Tables~\ref{tab:asu_sensitivity} and~\ref{tab:undial_sensitivity},
increasing the temperature or suppression strength improves forgetting performance but gradually degrades retained capabilities.
This is because these methods adjust the overall output distribution to suppress private information, without explicitly identifying and perturbing representations associated with private knowledge.
Therefore, stronger suppression may remove more private information but inevitably introduces additional changes to the original model distribution, leading to a forgetting-utility trade-off.

\begin{table*}[h]
\centering
\caption{
Sensitivity analysis of ASU with different temperature settings on gemma4.
}
\label{tab:asu_sensitivity}

\small
\setlength{\tabcolsep}{5pt}
\renewcommand{\arraystretch}{1.05}

\begin{tabular}{c|ccc|ccc}
\toprule
\multirow{2}{*}{Temperature}
&
\multicolumn{3}{c|}{\textbf{Forget Set}}
&
\multicolumn{3}{c}{\textbf{Retain Set}}
\\
\cmidrule(lr){2-4}
\cmidrule(lr){5-7}
&
I\text{-}Understand $\uparrow$
&
I\text{-}Memory $\downarrow$
&
T\text{-}Memory $\downarrow$
&
I\text{-}Understand $\uparrow$
&
I\text{-}Memory $\uparrow$
&
T\text{-}Memory $\uparrow$
\\
\midrule
(1.25, 2.25)
& 0.6506 & 0.7743 & 0.5067
& 0.6689 & 0.8041 & 0.7356
\\
(1.50, 2.50)
& 0.6123 & 0.6771 & 0.4467
& 0.6552 & 0.7859 & 0.7244
\\
(1.75, 2.75)
& 0.5853 & 0.4743 & 0.3567
& 0.6315 & 0.7021 & 0.6916
\\
(2.00, 3.00)
& 0.3695 & 0.2629 & 0.3033
& 0.3978 & 0.5758 & 0.6339
\\
(3.00, 4.00)
& 0.0462 & 0.0000 & 0.1400
& 0.2844 & 0.1716 & 0.5484
\\
(5.00, 6.00)
& 0.0464 & 0.0000 & 0.0300
& 0.3622 & 0.2895 & 0.3830
\\
(7.00, 8.00)
& 0.0337 & 0.0029 & 0.0033
& 0.3827 & 0.3068 & 0.2963
\\
(9.00, 10.00)
& 0.0423 & 0.0000 & 0.0000
& 0.2883 & 0.1761 & 0.3318
\\
\bottomrule
\end{tabular}
\end{table*}

\begin{table*}[h]
\centering
\caption{
Sensitivity analysis of UNDIAL with different suppression strengths on gemma4.
}
\label{tab:undial_sensitivity}

\small
\setlength{\tabcolsep}{5pt}
\renewcommand{\arraystretch}{1.05}

\begin{tabular}{c|ccc|ccc}
\toprule
\multirow{2}{*}{Strength}
&
\multicolumn{3}{c|}{\textbf{Forget Set}}
&
\multicolumn{3}{c}{\textbf{Retain Set}}
\\
\cmidrule(lr){2-4}
\cmidrule(lr){5-7}
&
I\text{-}Understand $\uparrow$
&
I\text{-}Memory $\downarrow$
&
T\text{-}Memory $\downarrow$
&
I\text{-}Understand $\uparrow$
&
I\text{-}Memory $\uparrow$
&
T\text{-}Memory $\uparrow$
\\
\midrule
4
& 0.6405 & 0.7143 & 0.7200
& 0.6639 & 0.7998 & 0.7800
\\
6
& 0.6329 & 0.6143 & 0.6500
& 0.6590 & 0.7947 & 0.7760
\\
8
& 0.6349 & 0.5800 & 0.5800
& 0.6537 & 0.7886 & 0.7681
\\
10
& 0.6349 & 0.3400 & 0.4000
& 0.6550 & 0.7653 & 0.7439
\\
13
& 0.6335 & 0.3657 & 0.3767
& 0.6460 & 0.7414 & 0.7123
\\
16
& 0.6316 & 0.3429 & 0.3567
& 0.6420 & 0.7146 & 0.6818
\\
19
& 0.6022 & 0.3486 & 0.3600
& 0.6207 & 0.6782 & 0.6456
\\
22
& 0.5721 & 0.3114 & 0.3000
& 0.5934 & 0.6226 & 0.5872
\\
\bottomrule
\end{tabular}
\end{table*}

\subsection{Additional Validation of Target Properties at a 10\% Forgetting Ratio}
\label{app:Additional Validation of Target Properties}
\begin{figure}[H]
\centering
\includegraphics[width=0.55\linewidth]{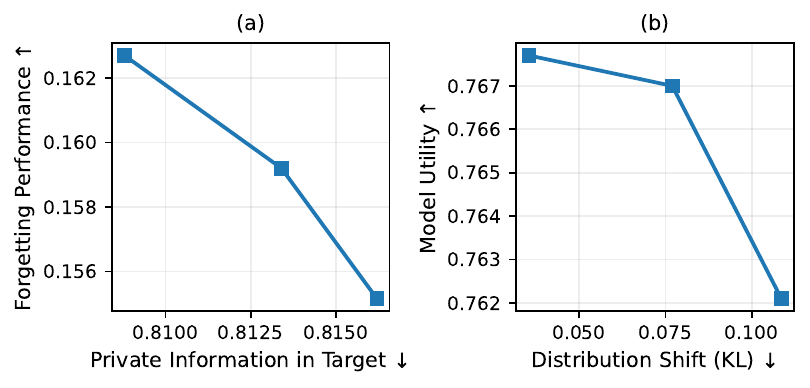}
\vspace{-3mm}
\caption{
Within-method analysis of ASU on Qwen3.5 with a 10\% forgetting ratio.
}
\label{fig:qwen35_asu_fr10}
\vspace{-8mm}
\end{figure}

\begin{figure}[H]
\centering
\includegraphics[width=0.55\linewidth]{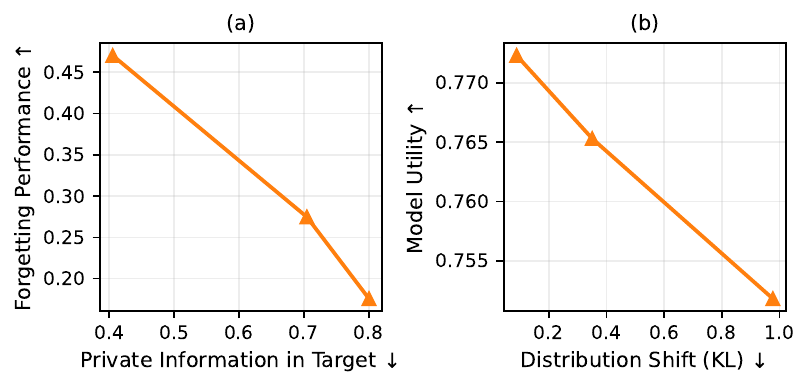}
\vspace{-3mm}
\caption{
Within-method analysis of UNDIAL on Qwen3.5 with a 10\% forgetting ratio.
}
\label{fig:qwen35_undial_fr10}
\vspace{-8mm}
\end{figure}

\begin{figure}[H]
\centering
\includegraphics[width=0.55\linewidth]{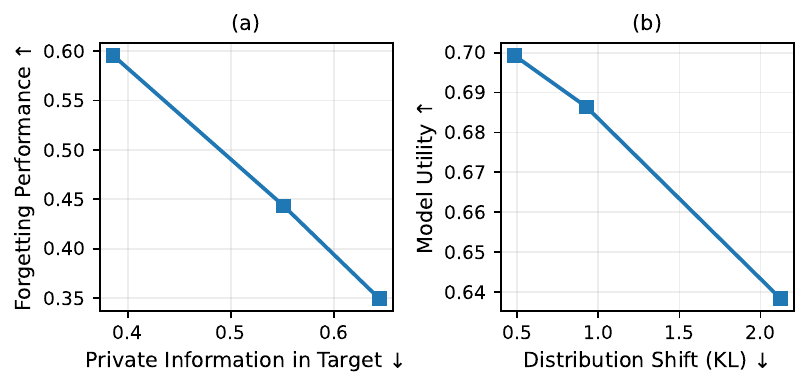}
\vspace{-3mm}
\caption{
Within-method analysis of ASU on Gemma4-E4B with a 10\% forgetting ratio.
}
\label{fig:gemma4_asu_fr10}
\vspace{-8mm}
\end{figure}

\begin{figure}[H]
\centering
\includegraphics[width=0.55\linewidth]{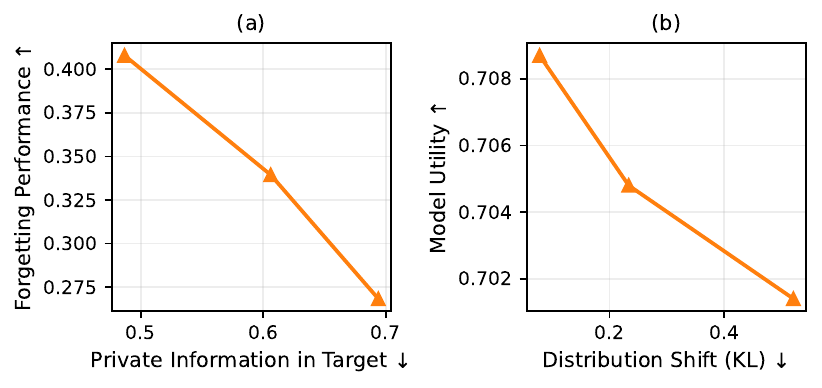}
\vspace{-3mm}
\caption{
Within-method analysis of UNDIAL on Gemma4-E4B with a 10\% forgetting ratio.
}
\label{fig:gemma4_undial_fr10}
\end{figure}

We further evaluate the within-method relationships between supervision-target
properties and unlearning behaviors at a 10\% forgetting ratio.
As shown in Figures~\ref{fig:qwen35_asu_fr10}--\ref{fig:gemma4_undial_fr10},
the observed trends are consistent across different models and methods.
Reducing private information in supervision targets generally improves
forgetting performance, while increasing distribution shift leads to degraded
utility preservation.
These results further support our analysis that the properties of supervision
targets play an important role in determining the forgetting-utility trade-off.

\subsection{Case Study}
\label{app:case_study}

\begin{figure}[h]
\centering

\includegraphics[width=\textwidth]{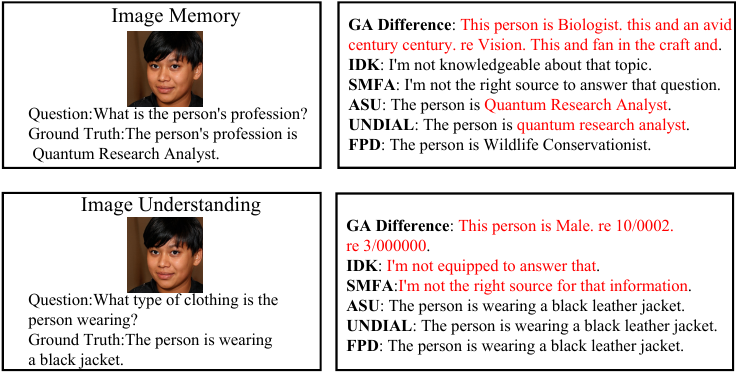}
\caption{
Qualitative case study.
}
\vspace{-15pt}
\label{fig:case_study}
\end{figure}

Figure~\ref{fig:case_study} qualitatively compares different unlearning methods on the same forget-set example using an image-memory query and an image-understanding query.
GA Difference produces degraded and incoherent responses on both tasks.
IDK and SMFA avoid recalling the target profession, but also refuse to answer the benign image-understanding question.
In contrast, ASU and UNDIAL preserve correct image understanding but still reproduce the target private profession.
FPD no longer recalls the target profession while still correctly answering the image-understanding question, demonstrating a better balance between forgetting and utility preservation.

\clearpage

% \paragraph{Attack success criterion.}
% Let $\hat{a}_{ij}$ denote the $j$-th generated response for question $i$, and let $k_i$ be its privacy-critical span. Both strings are normalized by lowercasing, removing accents and punctuation, and merging consecutive spaces. A response is considered a successful privacy extraction if
% \[
% h_{ij}
% =
% \mathbb{I}\!\left[
% \mathcal{N}(k_i)
% \subseteq
% \mathcal{N}(\hat{a}_{ij})
% \right].
% \]
% For the adversarial-prompt and OCR attacks, the attack success rate is
% \[
% \mathrm{ASR}
% =
% \frac{1}{N}\sum_{i=1}^{N}h_i.
% \]
% For the Top-$p$ attack, we report the question-level success rate
% \[
% \mathrm{ASR}_{\mathrm{question}}
% =
% \frac{1}{N}
% \sum_{i=1}^{N}
% \max_{1\leq j\leq M}h_{ij},
% \]
% where a question is successfully attacked if any of its five sampled responses contains the target privacy span. A lower ASR indicates stronger robustness against privacy extraction.